\documentclass[runningheads]{llncs}
\usepackage{algorithm}
\usepackage{algpseudocode}
\usepackage{multirow}

\usepackage{amssymb}
\usepackage{caption}
\usepackage{subcaption}
\usepackage{xcolor,colortbl}
\definecolor{Gray}{gray}{0.85}
\newcolumntype{a}{>{\columncolor{Gray}}l}
\usepackage[T1]{fontenc}
\usepackage{graphicx}
\def\Fig#1{{Fig.~\ref{fig:#1}}}

\def\Table#1{{Table~\ref{tbl:#1}}}

\begin{document}
\title{Anatomy-Aware Lymph Node Detection in \\Chest CT using Implicit Station Stratification}
\titlerunning{Lymph Node Detection using Station Stratification}
%
\author{Ke Yan\inst{1,2} \and 
Dakai Jin\inst{1} \and
Dazhou Guo\inst{1} \and
Minfeng Xu\inst{1,2} \and
Na Shen\inst{3} \and \\
Xian-Sheng Hua\inst{1,2} \and
Xianghua Ye\inst{4} \and
Le Lu\inst{1} }

\institute{DAMO Academy, Alibaba Group \and
Hupan Lab, 310023, Hangzhou, China \and
Zhongshan Hospital of Fudan University, Shanghai, China \and
The First Affiliated Hospital Zhejiang University, Hangzhou, China \\
yanke.yan@alibaba-inc.com
}
\authorrunning{K. Yan et al.}
%

%
\maketitle              
\begin{abstract}
Finding abnormal lymph nodes in radiological images is highly important for various medical tasks such as cancer metastasis staging and radiotherapy planning. Lymph nodes (LNs) are small glands scattered throughout the body. They are grouped or defined to various LN stations according to their anatomical locations. The CT imaging appearance and context of LNs in different stations vary significantly, posing challenges for automated detection, especially for pathological LNs. Motivated by this observation, we propose a novel end-to-end framework to improve LN detection performance by leveraging their station information. We design a multi-head detector and make each head focus on differentiating the LN and non-LN structures of certain stations. Pseudo station labels are generated by an LN station classifier as a form of multi-task learning during training, so we do not need another explicit LN station prediction model during inference. Our algorithm is evaluated on 82 patients with lung cancer and 91 patients with esophageal cancer. The proposed implicit station stratification method improves the detection sensitivity of thoracic lymph nodes from 65.1\% to 71.4\% and from 80.3\% to 85.5\% at 2 false positives per patient on the two datasets, respectively, which significantly outperforms various existing state-of-the-art baseline techniques such as nnUNet, nnDetection and LENS.

\keywords{Lymph node detection \and Lymph node station \and CT.}
\end{abstract}
\section{Introduction}
Lymph nodes play essential roles in the staging and treatment planning of general cancer patients~\cite{Eisenhauer2009RECIST,Mao2014current}. As cancer evolves, tumor cells can spread to lymph nodes and cause them to metastasize and possibly enlarge. Finding all of the abnormal (metastatic) lymph nodes is a crucial task for radiologists and oncologists. Computed tomography (CT) is the primary modality for tumor imaging in the chest~\cite{Sharma2004pattern}. In CT, most lymph nodes can be identified as small, oval-shaped structures with soft-tissue intensity, which are challenging to be differentiated from surrounding soft tissues such as vessels, esophagus, and muscles. Due to its importance and difficulty, automatic lymph node (LN) detection and segmentation has been attracting increasing attentions~\cite{Feulner2013LN,Roth201425D,Oda2018dense,Bouget2019semantic,Zhu2020gating,Iuga2021automated}. Convolutional neural network (CNN) is becoming the mainstream method in recent years. Oda et al.~\cite{Oda2018dense} trained a 3D U-Net using not only LN annotations but also neighboring organs to reduce oversegmentation of LNs. Bouget et~al.~\cite{Bouget2019semantic} combined the outputs of 2D U-Net and Mask R-CNN to predict both LNs and neighboring organs. Yan et al.~\cite{Yan2020LENS} showed that jointly learning multiple datasets improved LN detection accuracy. Zhu et al.~\cite{Zhu2020gating} divided LNs into two subclasses of tumor-proximal and tumor-distal ones and used a U-Net with two decoder branches to learn the two groups separately. Iuga et al.~\cite{Iuga2021automated} designed a neural network with multi-scale inputs to fuse information from multiple spatial resolutions.

\begin{figure}[t]
\centering
\includegraphics[width=\textwidth, trim=0 230 50 0, clip]{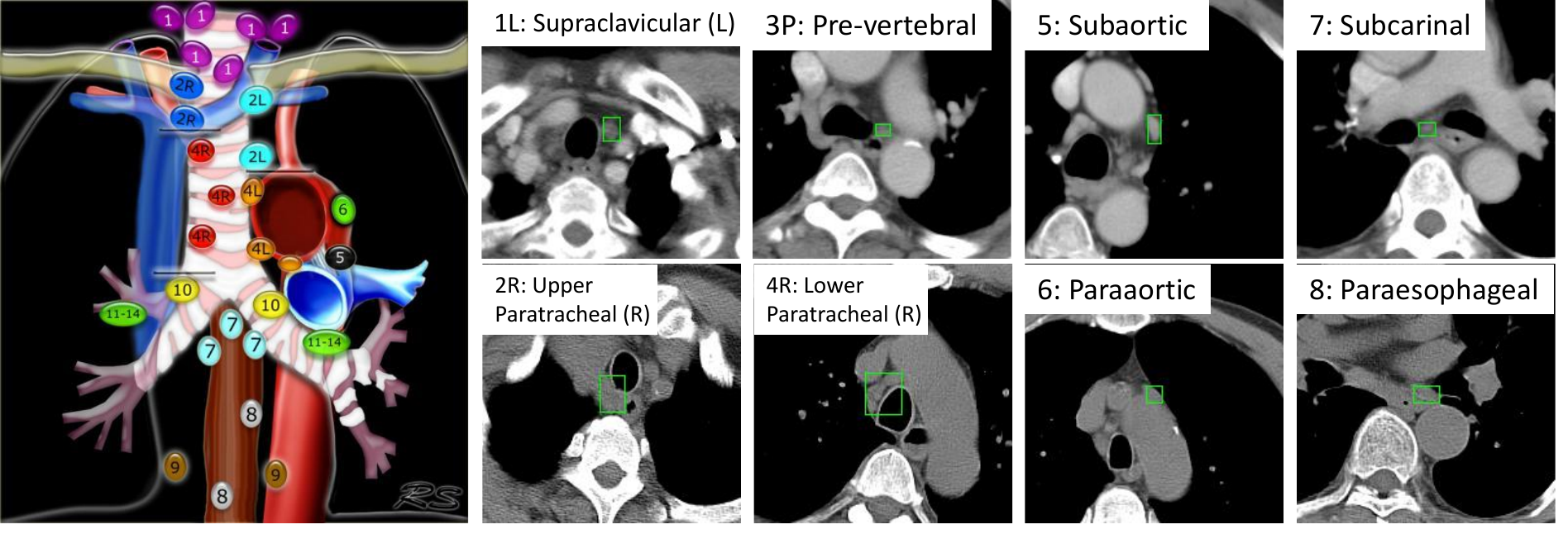}  
\caption{Lymph node (LN) stations defined for lung cancer staging~\cite{Sherief2014IASLC}. The anatomical map on the left is reproduced from~\cite{Smithuis2010}. LN examples in some stations are shown on the right in green boxes, either in contrast-enhanced (1st row) or non-contrast (2nd row) images. Note the significant diversity of appearance across stations.} \label{fig:station_intro}
\end{figure}

Different from other types of lesions (e.g., lung nodules) that typically locate in one organ, LNs scatter throughout the body. The anatomical location of a metastatic lymph node is an important indicator to determine the stage of the cancer and even the subsequent treatment recommendations. Taking lung cancer as an example, the International Association for the Study of Lung Cancer (IASLC) defined 14 lymph node stations in the chest based on their relative position with adjacent organs~\cite{Sherief2014IASLC}, as shown in \Fig{station_intro}. 
We can observe that LNs in different stations are surrounded by varying organs, thus show very diverse contextual layouts. To detect an LN is essentially to distinguish it from surrounding confounding organs and structures, therefore, detecting LNs in different stations may actually be considered as different tasks. Most existing works treat LNs in all stations as one positive class and define other organs as one negative class. We would argue that this representation is suboptimal because the inter-class difference between LNs and non-LNs is sometimes very subtle (e.g., \Fig{station_intro} 2R and 8). If we mix the samples from all stations, the model may struggle to learn the coherent imaging feature of LNs and be distracted by their contextual appearance. In this work, we propose to first stratify LNs and non-LNs based on their stations, and then train an LN vs.~non-LN classifier for each station group. \Fig{intuition} illustrates our intuition. In addition, the distributions of shape and size of LNs vary in different stations~\cite{Mao2014current,Sharma2004pattern}. Our stratification strategy could also handle this variation better by separately modeling each station.

In this paper, we instantiate this strategy and propose a station-stratified LN detector. It is based on the widely-used two-stage CNN detection architecture~\cite{Ren2015faster,Yan2020LENS} with a novel detection branch and a station branch simultaneously. The detection branch contains multiple output heads, each focusing on classifying LN/non-LN in one station group. The station branch predicts a probability vector for each proposal indicating its station group, which in turn is used by the detection branch to compute a weighted loss in training and a final LN likelihood in inference. The group can either be stations or super-stations (by grouping similar stations). A related but different method is~\cite{Zhu2020gating}. They proposed a segmentation method that groups LNs according to their distance with the tumor, thus the location of tumor needs to be known in prior. We stratify LNs according to the anatomy-related stations and no tumor location is needed. Our method is more widely applicable even for non-cancer patients as a form of screening abnormal LNs by stations implicitly. No extra cost on LN station segmentation is needed in inference. While LN groups in~\cite{Zhu2020gating} are manually computed and the distance threshold needs to be tuned, ours are predicted by a station branch automatically. Our algorithm employs a 2.5D backbone for better efficiency and accuracy. To convert the predicted 2D boxes to 3D ones, we further design a novel lesion-centric box stacking and merging algorithm. 

The proposed framework is extensively evaluated on two datasets of 82 patients with lung cancer and 91 patients with esophageal cancer. A total of 1,380 lymph nodes were annotated in the 14 IASLC stations. By employing the proposed station stratification strategy alone, our LN detection sensitivity is improved from 65.1\% to 71.4\% and from 80.3\% to 85.5\% at 2 false positives (FPs) per patient in the two datasets, respectively, outperforming various strong mainstream methods such as nnUNet~\cite{Isensee2021nnunet}, nnDetection~\cite{Baumgartner2021nndet}, and LENS~\cite{Yan2020LENS}. To the best of our knowledge, we are the first to demonstrate that the station information can be used to improve LN detection effectively (from recent literature reported). While most prior studies used contrast-enhanced (CE) CTs, we also run our method on 85 more challenging non-contrast (NC) CT scans. Joint learning of CE and NC CT imaging modalities achieves a sensitivity of 83.8\% at 2 FPs per patient of NC CT scan, which is an encouraging result for scenarios such as lung nodule screening and radiotherapy planning~\cite{Zhu2020gating}.

\section{Method}

\begin{figure}[h!]
\centering
\includegraphics[width=\textwidth, width=\columnwidth, trim=0 320 230 0, clip]{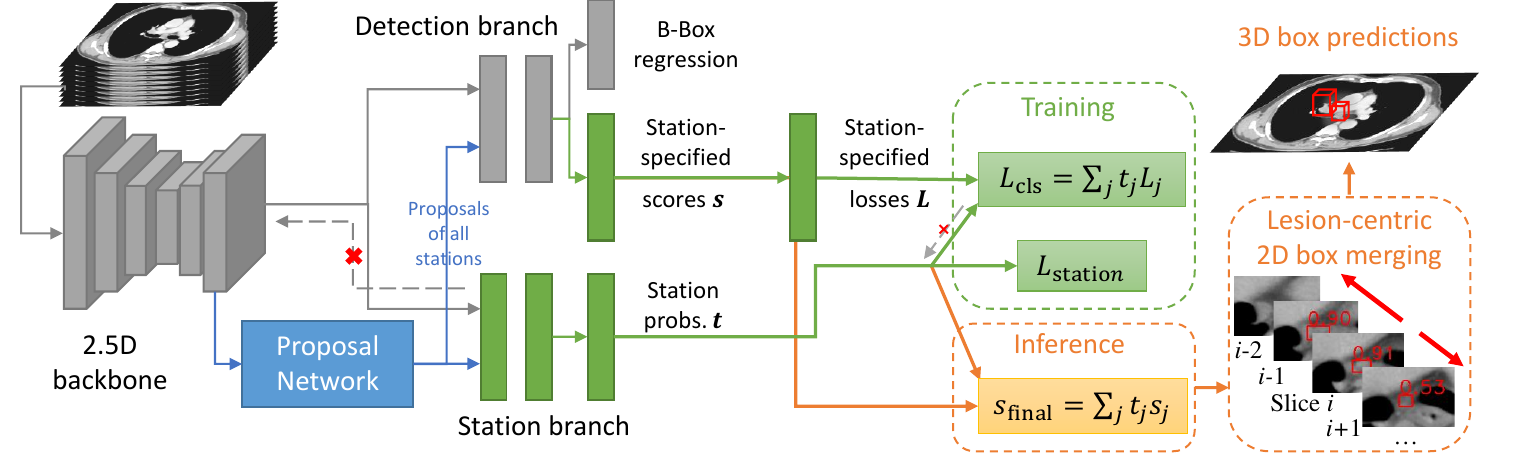}  
\caption{Framework of our proposed station-stratified LN detector. The blocks in green and orange are our key technical novelties. Red cross mark means gradient stopping.} \label{fig:framework} 
\end{figure}

\begin{figure}[h!]
\centering
\includegraphics[width=\textwidth, width=\columnwidth, trim=0 270 160 0, clip]{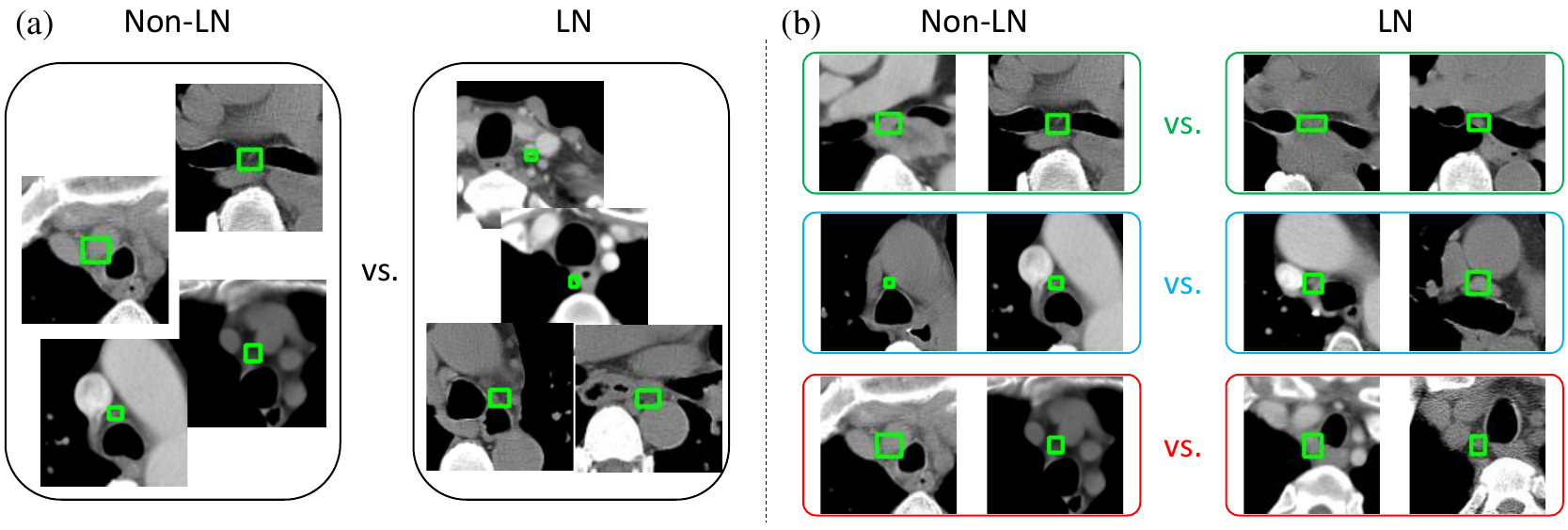}  
\caption{(a) Existing LN detection algorithms mix samples in different stations and train one classifier. (b) Our method stratifies samples based on stations and learns station-specific classifiers. 
Samples in each group share similar contextual appearance, so the model can focus on mining the subtle discriminative features to separate LNs/non-LNs.} \label{fig:intuition} 
\end{figure}

The framework of our proposed method is illustrated in \Fig{framework}. It is based on the widely-used two-stage detection framework Faster R-CNN~\cite{Ren2015faster}. 
The input of the network is multiple consecutive axial CT slices. We adopt the 2.5D design in MULAN~\cite{Yan2019MULAN} as backbone. It extracts 2D features for the CT slices and aggregates them to fuse 3D context information, which is important for distinguishing LNs from other tube-shaped organs such as vessels and esophagus. We empirically find the 2.5D network outperforms pure 3D ones in both convergence speed and accuracy. The fused feature map is fed to a proposal network such as the Fully Convolutional One-Stage detector (FCOS)~\cite{Tian2019FCOS}. It learns to generate 2D LN proposals in all LN stations. We observed that the vast majority of proposals concentrate in LN station areas, but some of them are confounding false positives (FPs) such as vessels, connective tissue, and esophageal tumor inside the station, see examples in the non-LN columns of \Fig{intuition}. This indicates that the proposal network has successfully learned the image context of the LNs, but struggles to differentiate some subtle structures inside the stations.

To solve this problem, we force the network to disambiguate true positive (TPs) and FPs inside the same station to learn more discriminative features. Specifically, we design a novel multi-head detection branch and a station branch as the second stage of the detector. Suppose that there are $c$ stations, the {\bf station branch} predicts a probability vector $\mathbf{t}_i\in \mathbb{R}^c$ to classify the station of each proposal $i$. It is trained on manually annotated station labels using the cross-entropy loss. The loss is only computed on TP proposals since only TPs have station labels, but the branch can predict station probabilities for both TPs and FPs. Similar to Faster R-CNN~\cite{Ren2015faster}, the {\bf detection branch} contains a classification layer and a bounding-box regression layer. In our algorithm, we aim to train $c$ LN vs.~non-LN classifiers, each corresponding to one station (or station group). Therefore, the classification layer will output a station-specified score vector $\mathbf{s}_i\in \mathbb{R}^c$ for each proposal $i$. These scores have a common LN/non-LN label $y_i$, so we can compute a binary cross-entropy loss for each score, forming a station-specified loss vector $\mathbf{L}_i\in \mathbb{R}^c$. Finally, we use the station probabilities $\mathbf{t}_i$ to compute a weighted sum of them:
\begin{equation}
    L_{\mathrm{cls}} = \frac{1}{n}\sum_{i=1}^n \sum_{j=1}^c t_{ij} \left(y_i \log\sigma(s_{ij}) + (1-y_i)(1-\log\sigma(s_{ij}))\right),
\end{equation}
where $n$ is the number of proposals in a mini-batch, $\sigma$ is the sigmoid function. During inference, station label is no longer needed because the station branch has learned to predict it. We use the predicted $\mathbf{t}$ to compute a weighted score $s_{i, \mathrm{final}}=\sum_{j=1}^c t_{ij} s_{ij}$ for each proposal $i$.

Ideally, $\mathbf{t}_{ij}$ should be 1 if proposal $i$ belongs to station $j$ and 0 otherwise, so only $L_{ij}$ will be counted in $L_\mathrm{cls}$, making classifier $j$ receive positive and negative samples in the station $j$ alone. However, some proposals may lie in the intersection area of multiple stations. The predicted station probabilities are also not ideal. Thus, we use a soft-gated loss $L_\mathrm{cls}$ weighted by $\mathbf{t}$, which is more robust than hard-gating each proposal to only one classifier. The $c$ classifiers are all built upon the feature vector of the final fully-connected (FC) layer in the detection branch, which can be viewed as finding an optimal subspace for each station in the feature space. As shown in \Fig{framework}, the station branch does not back-propagate gradients to the backbone
. We find it yield better detection performance, possibly because it can make the backbone focus on learning features for LN vs.~non-LN. In this study, our goal is not improving the station classification accuracy. The mean area-under-ROC-curve (AUC) for station classification is 93.5\% in this setting, showing that station classification is a relatively simple task.

The proposed method predicts 2D boxes for each CT slice. It is necessary to merge 2D boxes to 3D ones to describe 3D lesions. LENS~\cite{Yan2020LENS} proposed a merging algorithm. It starts from the boxes in the first slice, and then merges boxes in the second slice that overlaps with those in the first slice in the axial plane, and repeats until the last slice. This algorithm has a drawback: it starts to generate each 3D box from its first 2D box, which corresponds to the top edge of a lesion that may be inaccurate in detection with a low confidence score. Inspired by the non-maximum suppression (NMS) algorithm, we propose to start generating each 3D box from its 2D box with the highest confidence score, as detailed in Algorithm \ref{algo:merge} and \Fig{framework}. Our experiments show that this lesion-centric merging strategy outperforms the slice-wise scheme in~\cite{Yan2020LENS}.

\begin{algorithm}[]
	\small
	\caption{3D box generation by lesion-centric 2D box merging}\label{algo:merge}
	\begin{algorithmic}[1]
		\Require A list of predicted 2D boxes $B_2$; Intersection-over-union (IoU) threshold $\theta$.
		\Ensure A list of merged 3D boxes $B_3$.
		\While {$B_2$ is not empty}
		\State Take $b\in B_2$ with the highest confidence score. Suppose $b$ is on slice $i$.
		\State Create a new 2D box list $T=\{b\}$
		\For {slices $i+1, i+2, \cdots$}
		\If {$\exists\, \tilde{b}\in B_2, \mathrm{IoU}(b, \tilde{b})>\theta$} {$T=T\cup\{\tilde{b}\}, B_2=B_2-\tilde{b}$} {~\bf else} stop iteration.
		\EndIf
		\EndFor
		\State Repeat steps 4--7 for slices $i-1, i-2, \cdots$
		\State Compute a 3D box $\hat{b}$ from $T$, whose $x,y,z$ ranges and confidence score is the maximum of the 2D boxes in $T$. $B_3=B_3\cup\{\hat{b}\}$
		\EndWhile
	\end{algorithmic}
\end{algorithm} 

\begin{figure}
     \centering
     \begin{subfigure}[b]{0.5\textwidth}
         \centering
         \includegraphics[width=\textwidth]{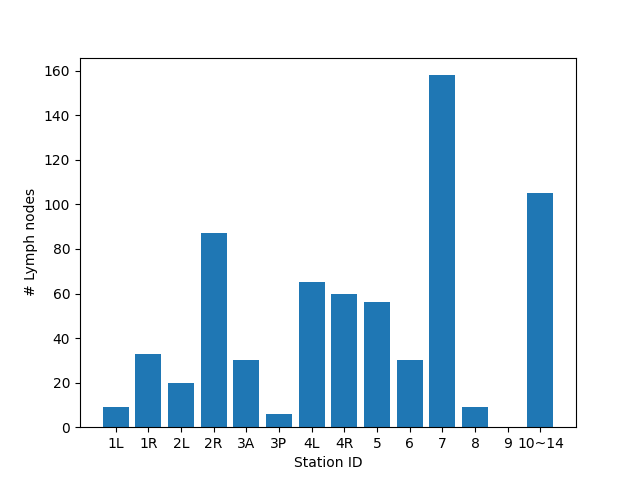}
        \caption{}
     \end{subfigure}
     \hspace{-3mm}
     \begin{subfigure}[b]{0.5\textwidth}
         \centering
         \includegraphics[width=\textwidth]{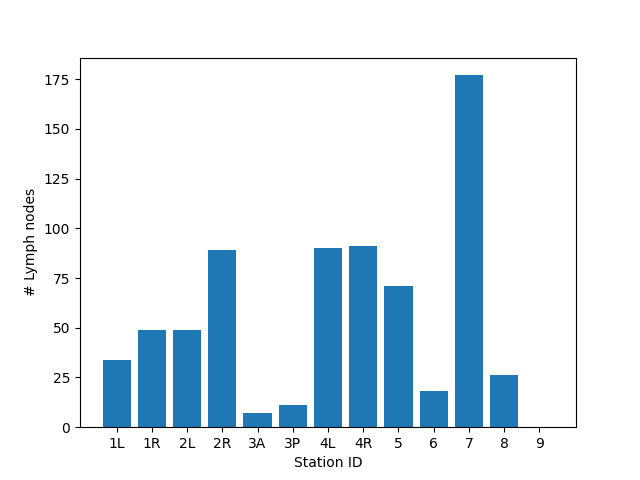}
         \caption{}
     \end{subfigure}
    \\
     \begin{subfigure}[b]{0.5\textwidth}
         \centering
         \includegraphics[width=\textwidth]{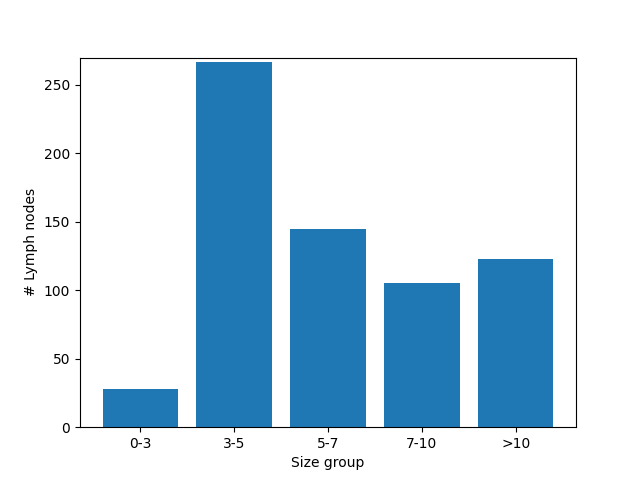}
         \caption{}
     \end{subfigure}
    \hspace{-3mm}
     \begin{subfigure}[b]{0.5\textwidth}
         \centering
         \includegraphics[width=\textwidth]{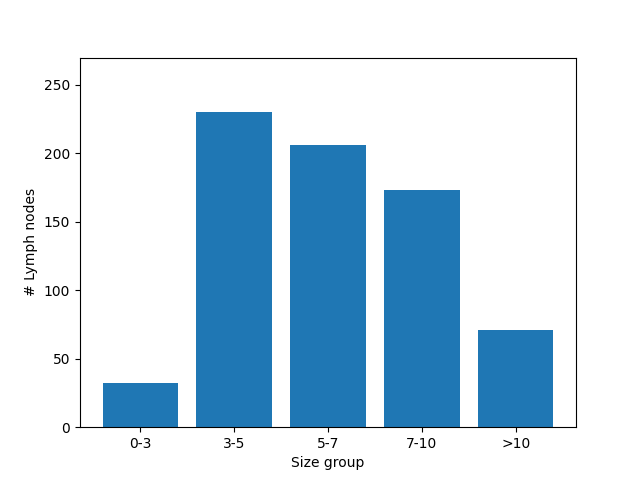}
         \caption{}
     \end{subfigure}
        \caption{(a) Station distribution of the lung cancer dataset. (a) Station distribution of the esophageal cancer dataset. (c) Size distribution (in mm) of the lung cancer dataset. (a) Size distribution (in mm) of the esophageal cancer dataset. }
        \label{fig:ds_dist}
\end{figure}
\section{Experiment}

\noindent{\bf Datasets.} Thoracic LNs can be affected by multiple cancer types~\cite{Sharma2004pattern}. In this work, we collected two datasets of different cancer origins. The {\bf lung cancer} dataset includes contrast-enhanced (CE) CTs of 82 patients. 668 LNs were annotated by three board-certified radiation oncologist with more than 10 years of experience. All visible LNs were comprehensively annotated, whose average long and short diameters~\cite{Eisenhauer2009RECIST} are $12.3\times7.4$mm (min.~1.5mm, max.~60.6mm). The {\bf esophageal cancer} dataset contains both CE and non-contrast (NC) CTs of 91 patients. 712 LNs in stations 1--9 with average diameters of $11.0\times 6.5$mm (min.~2.1mm, max.~27.0mm) were annotated by the same group of oncologists. The LNs were annotated on CE CTs in which they were more distinguishable. Then, we registered NC CTs to CE ones for each patient using DEEDS~\cite{Heinrich2012DEEDS}, followed by manual verification of the registration quality. In this way, we can train and evaluate our LN detector on NC CTs as well. The masks of LN stations 1--9 were also annotated in this dataset, from which we can infer the station label of each LN. We also trained an LN station segmentation algorithm~\cite{Guo2021station} using these annotations and applied it to the lung cancer dataset to infer their LN stations. Note that LNs in stations 10--14 (pulmonary nodes~\cite{Sherief2014IASLC}) exist in the lung cancer dataset but not in the esophageal cancer one. The station segmentation algorithm cannot predict stations 10--14. Hence, when applying it on the lung cancer dataset, we regarded all LNs outside its predicted masks as belonging to stations 10--14. See \Fig{ds_dist} for details about distribution of LN stations and sizes in the datasets.

\noindent{\bf Implementation details.} We implemented our algorithm using PyTorch 1.10 and mmDetection 2.18~\cite{mmdetection}. CT images were normalized using a spacing of $0.8\times 0.8 \times 2$mm and an intensity window of $[-200, 300]$ Hounsfield unit. Data augmentation included random scaling (0.7--1.4), cropping, rotation ($\pm 15^{\circ}$), intensity scaling (0.7--1.3), and gamma augmentation (0.7--1.5)~\cite{Perez-Garcia2020torchio}. In training, each mini-batch consisted of 4 samples, where each sample included 9 CT slices for 3D feature fusion~\cite{Yan2019MULAN}. The station branch had two 512D FC layers, whereas the detection branch had two 2048D FC layers. We used RAdam~\cite{Liu2020radam} to train for 10 epochs and set the base learning rate to 0.0001, and then reduced it by a factor of 10 after the 7th epoch. In each epoch, we used all positive slices (with LN annotations) and randomly sampled 2 times of negative slices (without annotations)~\cite{Yan2020LENS}. The entire training process took 1.5h for the esophageal cancer dataset on a Tesla V100 GPU. In the 2D box merging algorithm, we set the IoU threshold $\theta$ to 0.7.

\noindent{\bf Evaluation metrics.} For both datasets, we randomly split the data into 60\% training, 15\% validation, and 25\% testing in the patient level. For the esophageal cancer dataset, we trained a joint model for CE and NC images and show their performance in the test set separately. Following previous lesion detection works~\cite{Yan2020LENS,Bouget2019semantic,Liu2016station,Yan2019MULAN}, we use the free-response receiver operating characteristic (FROC) curve as the evaluation metric and report the sensitivity at different FP levels. When comparing each detected 3D box with the ground-truth 3D boxes, if the 3D intersection over detected bounding-box ratio (IoBB) is larger than 0.3, the detected box is counted as hit~\cite{Yan2020LENS}. According to the RECIST guideline~\cite{Eisenhauer2009RECIST}, LNs with short axis less than 10mm are considered normal. However, some studies~\cite{Sharma2004pattern} show that metastatic LNs can be smaller than 10mm. Therefore, we set a smaller size threshold and aim to detect LNs larger than 7mm during inference. If a ground-truth LN smaller than 7mm is detected, it is neither counted as a TP nor an FP. In training, we still use all LN annotations.

\begin{table}[t]
\caption{Sensitivity (\%) at 0.5, 1, 2, and 4 FPs per patient on the lung and esophageal cancer datasets. The number of heads $c$ is varied, where $c=1$ is the baseline.}\label{tbl:res_diffc}
\setlength{\tabcolsep}{3pt}
\renewcommand{\arraystretch}{1.2}
\centering
\begin{tabular}{l|ccccc|ccccc|ccccc}
\hline
& \multicolumn{5}{c|}{\emph{Lung}}    & \multicolumn{5}{c|}{\emph{Esophageal CE}} & \multicolumn{5}{c}{\emph{Esophageal NC}} \\
\cline{2-16}
$c$	& 0.5	& 1	& 2	& 4	& Avg.	& 0.5	& 1	& 2	& 4	& Avg.	& 0.5	& 1	& 2	& 4	& Avg.	\\
\hline
1	& 52 	& 60 	& 65 	& 70 	& 61.9  & \bf 66 	& 70 	& 80 	& 87 	& 75.7 	& 59 	& 68 	& 81 	& 88 	& 73.9 \\
6	& \bf 60 	& \bf 68 	& \bf 71 	& \bf 76 & \bf 69.0$_{~7.1\scriptscriptstyle\uparrow}$ & 60 	& 76 	& \bf 86 	& 90 	& \bf 77.8$_{~2.1\scriptscriptstyle\uparrow}$ 	& \bf 63 	& 71 	& 79 	&\bf  91 	& 76.1$_{~2.2\scriptscriptstyle\uparrow}$	 \\
8	& 56 	& 64 	& \bf 71 	& 75 	& 66.3$_{~4.4\scriptscriptstyle\uparrow}$ & 58 	& 72 	& 84 	& \bf 92 	& 76.6$_{~0.9\scriptscriptstyle\uparrow}$ 	& 57 	& \bf 75 	& 82 	& 88 	& 75.7$_{~1.8\scriptscriptstyle\uparrow}$	  \\
14	& 49 	& 62 	& 68 	& \bf 76 	& 63.9$_{~2.0\scriptscriptstyle\uparrow}$ 	& 63 	& \bf 78 	& 83 	& 86 	& 77.3$_{~1.6\scriptscriptstyle\uparrow}$ 	& \bf 63 	& 72 	& \bf 84 	& 88 	& \bf 76.8$_{~2.9\scriptscriptstyle\uparrow}$  \\
\hline
\end{tabular} 
\end{table}

\noindent{\bf Quantitative results.} First, we validate our key assumption: stratification of samples based on LN stations improves detection accuracy. Results on the two datasets are displayed in \Table{res_diffc}. $c=1$ means no stratification; $c=14$ means the most fine-grained stratification. We also tried to group some stations to super-stations according to radiological literature~\cite{Sherief2014IASLC}, resulting in $c=6$ or 8. 
Note that the lung dataset has one more station label (pulmonary nodes) than the esophageal dataset, so the actual $c$ used for the latter dataset is 1, 5, 7, and 13. In \Table{res_diffc}, station stratification consistently improves accuracy. Increasing $c$ enhances the purity of each group but also reduces the number of samples in each classifier, which is the possible reason why $c=6$ achieves the most significant improvement in the lung dataset. In the following experiments, we will use $c=6$.
The 6 groups are~\cite{Sherief2014IASLC}: supraclavicular (stations 1L, 1R), superior mediastinal (2L, 2R, 3A, 3P, 4L, 4R), aortopulmonary (5, 6), subcarinal (7), inferior mediastinal (8, 9), and pulmonary (10--14) nodes. Detection performance with different size thresholds is shown in \Table{size_th}.

\begin{table}[]
\caption{Comparison of sensitivity (\%) at different FPs per patient on each dataset. The performance of lymph nodes with different sizes are reported based on their short axis diameters.}
\label{tbl:size_th}
\setlength{\tabcolsep}{4pt}
\renewcommand{\arraystretch}{1.2}
\centering
\begin{tabular}{l|lllla|lllla|lllla}
\hline
& \multicolumn{5}{c|}{\emph{Lung}} & \multicolumn{5}{c|}{\emph{Esophageal CE}} & \multicolumn{5}{c}{\emph{Esophageal NC}}\\
\cline{2-16}
Size &	0.5	& 1	& 2	& 4	& Avg.	& 0.5	& 1	& 2	& 4	& Avg.	& 0.5	& 1	& 2	& 4	& Avg. \\
\hline
All	& 34 	& 41 	& 51 	& 59 	& 47.1 	& 38 	& 52 	& 59 	& 64 	& 53.4 	& 39 	& 45 	& 52 	& 63 	& 49.9 \\
> 5mm	& 52 	& 60 	& 63 	& 70 	& 61.1 	& 50 	& 66 	& 73 	& 79 	& 67.1 	& 50 	& 59 	& 67 	& 79 	& 63.6  \\
> 7mm	& 60 	& 68 	& 71 	& 76 	& 69.0 	& 60 	& 76 	& 86 	& 89 	& 77.8 	& 63 	& 71 	& 79 	& 91 	& 76.1 \\
> 10mm	& 74 	& 76 	& 79 	& 85 	& 78.7 	& 44 	& 78 	& 83 	& 89 	& 73.6 	& 64 	& 79 	& 93 	& 100 	& 83.9 \\
\hline
\end{tabular}
\end{table}

Next, we evaluate alternative strategies of our algorithm, see \Table{ablation}. We trained $c$ station-stratified classifiers. Another possibility is not to stratify samples using stations, but to use all samples to train each classifier and average their prediction during inference. This strategy did not bring improvement in \Table{ablation} row (b), showing the station information is useful and our performance gain is not simply due to increase of parameters and ensemble of predictions. One way to utilize station information is to train a $c$-way multi-class classifier, instead of the $c$ binary classifiers in our algorithm. In row (c), multi-class classification did not help. It asks the model to distinguish LNs of different stations, but LN detection actually requires the model to distinguish LNs and non-LNs in each station, which is effectively achieved by our strategy. In our algorithm, we use a soft gating strategy to combine classifiers in training and inference by weighted sum. It is better than the hard gating strategy~\cite{Zhu2020gating} in row (d) which only considers the classifier with the highest station score. In row (e), we show that our lesion-centric 2D box merging outperforms the slice-wise method in~\cite{Yan2020LENS}.

\begin{table}[t]
\caption{Sensitivity (\%) averaged at 0.5, 1, 2, and 4 FPs per patient using different strategies. Sta.: LN station information. LM: Lesion-centric 2D box merging.}\label{tbl:ablation}
\setlength{\tabcolsep}{3pt}
\renewcommand{\arraystretch}{1.2}
\centering
\begin{tabular}{l|cc|cccc}
\hline
Method	&	Sta. &	LM & Lung	& Eso.~CE	& Eso.~NC	& Average\\
\hline
(a) No stratification	& 	& \checkmark	& 61.9 	& 75.7 	& 73.9 	& 70.5 \\
(b) Uniform stratification	& 	& \checkmark	& 62.3 	& 74.0 	& 69.3 	& 68.5 \\
(c) Multi-class	& \checkmark	& \checkmark	& 58.1 	& 74.7 	& 72.6 	& 68.5 \\
(d) Hard gating	& \checkmark	& \checkmark	& 64.7 	& \bf 79.9 	& 73.5 	& 72.7 \\
(e) Slice-wise 2D box merging~\cite{Yan2020LENS}	& \checkmark	& 	& 65.5 	& 75.0 	& 74.6 	& 71.7 \\
(f) Proposed	& \checkmark	& \checkmark	& \bf 69.0 	& 77.8 	& \bf 76.1 	& \bf  74.3 \\
\hline
\end{tabular}
\end{table}

\begin{table}[t]
\caption{Comparison of sensitivity (\%) at different FPs per patient on each dataset. nnUNet is a segmentor thus only has one FP point. 
The number in parenthesis in the last row represents the sensitivity of the proposed method at the FP point of nnUNet.}\label{tbl:diff_mtd}
\scriptsize
\setlength{\tabcolsep}{3pt}
\renewcommand{\arraystretch}{1.2}
\centering
\begin{tabular}{l|lllla|lllla|lllla|l}
\hline
& \multicolumn{5}{c|}{\emph{Lung}} & \multicolumn{5}{c|}{\emph{Esophageal CE}} & \multicolumn{5}{c|
}{\emph{Esophageal NC}} & Time\\
\cline{2-16}
Method &	0.5	& 1	& 2	& 4	& Avg.	& 0.5	& 1	& 2	& 4	& Avg.	& 0.5	& 1	& 2	& 4	& Avg.	& (s) \\
\hline
nnDetection~\cite{Baumgartner2021nndet}	& 47 	& 57 	& 62 	& 70 	& 58.9 	& 43 	& 64 	& 72 	& 75 	& 63.8 	& 59 	& 63 	& 69 	& 72 	& 65.8 	& 86 \\
MULAN~\cite{Yan2019MULAN}	& 43 	& 57 	& 60 	& 73 	& 58.3 	& 51 	& 63 	& 75 	& 79 	& 67.1 	& 55 	& 68 	& 72 	& 81 	& 68.9 	& \bf 1.5 \\
LENS~\cite{Yan2020LENS}	& 58 	& 60 	& 67 	& 71 	& 64.1 	& \bf 64 	& \bf 79 	& 82 	& 84 	& 77.3 	& 60 	& \bf 74 	& \bf 79 	& 81 	& 73.5 	& 2.1 \\
Proposed	& \bf 60 	& \bf 68 	& \bf 71 	& \bf 76 	& \bf 69.0 	& 60 	& 76 	& \bf 86 	& \bf 89 	& \bf 77.8 	& \bf 63 	& 71 	& \bf 79 	& \bf 91 	& \bf 76.1 	& \bf 1.5 \\
\hline
nnUNet~\cite{Isensee2021nnunet} & \multicolumn{5}{c|}{71.4@2.8 FPs (vs. {\bf 73.0})} & \multicolumn{5}{c|}{78.9@3.7 FPs (vs. {\bf 89.5})} & \multicolumn{5}{c|
}{79.4@3.2 FPs (vs. {\bf 89.7})} & 53 \\
\hline
\end{tabular}
\end{table}

\begin{figure}[h!]
\centering
\includegraphics[width=\textwidth, trim=0 200 250 0, clip]{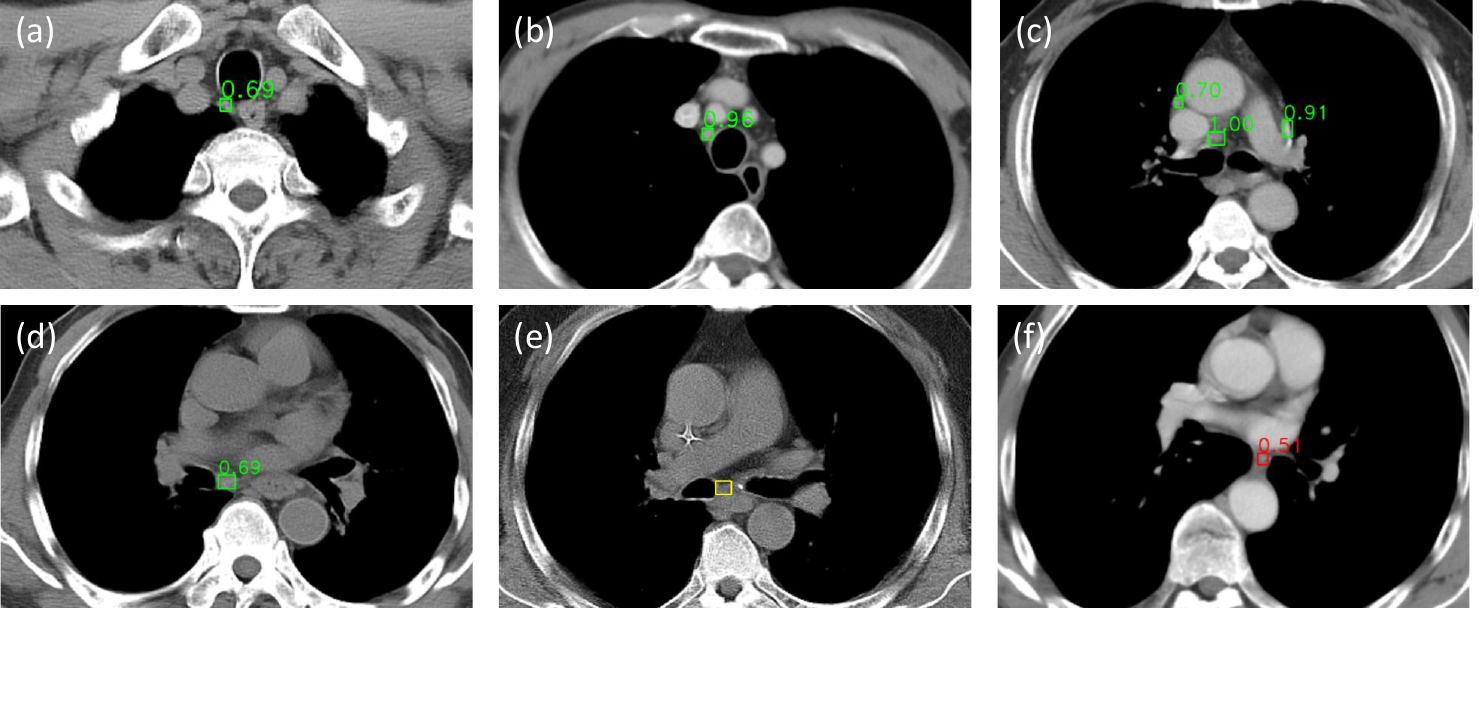}  
\caption{Exemplar detection results of our algorithm. LNs in different stations and image modalities (CE, NC) are shown. Green, yellow, and red boxes indicate TP, FN, and FPs, respectively, with the confidence score displayed above the box. In (a) and (b), our algorithm can differentiate between LNs and adjacent vessels and esophagus. (e) and (f) are failure cases. In (e), an LN in station 7 has indistinguishable intensity compared with surrounding tissue in an NC image, thus were missed by our algorithm. In (f), the esophagus was mistaken as an LN due to similar intensity and shape.}
\label{fig:examples}
\end{figure}

Finally, we compare our algorithm with prior works. nnDetection~\cite{Baumgartner2021nndet} is a self-configuring 3D detection framework utilizing test time augmentation and model ensemble. MULAN~\cite{Yan2019MULAN} is a 2.5D detection framework which learns lesion detection, classification, and segmentation in a multi-task fashion. LENS~\cite{Yan2020LENS} is the state of the art for 3D universal lesion detection. It improves lesion detection by jointly learning multiple datasets with a shared backbone and multiple proposal networks and detection heads. We trained it using both lung and esophageal datasets. nnUNet~\cite{Isensee2021nnunet} is a strong self-adapting framework that has been widely-used for medical image segmentation. Our proposed method achieves the best accuracy in all datasets with evident margins, while taking only 1.5s to infer a CT volume. 
It outperforms LENS even without multi-dataset joint training. More qualitative results are included in \Fig{examples}.

\section{Conclusion}
Lymph nodes (LNs) in different LN stations vary significantly in their contextual appearance. Inspired by this, we propose a lymph node detection algorithm that employs a station branch and a multi-head detection branch to train station-specialized classifiers. Our method is effective and efficient. It also significantly outperforms various leading lesion detection and segmentation methods ~\cite{Baumgartner2021nndet,Isensee2021nnunet,Yan2019MULAN,Yan2020LENS}, on two sets of patients with lung or esophageal cancers respectively. Our next step is to extend it to LNs in other body parts beyond thoracic CT scans.

%
%
%
\bibliographystyle{splncs04}
\bibliography{main}

\begin{thebibliography}{10}
\providecommand{\url}[1]{\texttt{#1}}
\providecommand{\urlprefix}{URL }
\providecommand{\doi}[1]{https://doi.org/#1}

\bibitem{Baumgartner2021nndet}
Baumgartner, M., Jaeger, P.F., Isensee, F., Maier-Hein, K.H.: {nnDetection: A
  Self-configuring Method for Medical Object Detection}. In: MICCAI (2021)

\bibitem{Bouget2019semantic}
Bouget, D., J{\o}rgensen, A., Kiss, G., Leira, H.O., Lang{\o}, T.: {Semantic
  segmentation and detection of mediastinal lymph nodes and anatomical
  structures in CT data for lung cancer staging}. International Journal of
  Computer Assisted Radiology and Surgery  (2019)

\bibitem{mmdetection}
Chen, K., Wang, J., Pang, J., Cao, Y., Xiong, Y., Li, X., Sun, S., Feng, W.,
  Liu, Z., Xu, J., Zhang, Z., Cheng, D., Zhu, C., Cheng, T., Zhao, Q., Li, B.,
  Lu, X., Zhu, R., Wu, Y., Dai, J., Wang, J., Shi, J., Ouyang, W., Loy, C.C.,
  Lin, D.: {MMDetection}: Open mmlab detection toolbox and benchmark. arXiv
  preprint arXiv:1906.07155  (2019)

\bibitem{Eisenhauer2009RECIST}
Eisenhauer, E.A., Therasse, P., Bogaerts, J., Schwartz, L.H., Sargent, D.,
  Ford, R., Dancey, J., Arbuck, S., Gwyther, S., Mooney, M., Rubinstein, L.,
  Shankar, L., Dodd, L., Kaplan, R., Lacombe, D., Verweij, J.: {New response
  evaluation criteria in solid tumours: Revised RECIST guideline (version
  1.1)}. European Journal of Cancer  \textbf{45}(2),  228--247 (2009)

\bibitem{Sherief2014IASLC}
El-Sherief, A.H., Lau, C.T., Wu, C.C., Drake, R.L., Abbott, G.F., Rice, T.W.:
  {International Association for the Study of Lung Cancer (IASLC) Lymph Node
  Map: Radiologic Review with CT Illustration}. RadioGraphics  \textbf{34}(6),
  1680--1691 (2014)

\bibitem{Feulner2013LN}
Feulner, J., {Kevin Zhou}, S., Hammon, M., Hornegger, J., Comaniciu, D.: {Lymph
  node detection and segmentation in chest CT data using discriminative
  learning and a spatial prior}. Medical Image Analysis  \textbf{17}(2),
  254--270 (2013)

\bibitem{Guo2021station}
Guo, D., Ye, X., Ge, J., Di, X., Lu, L., Huang, L., Xie, G., Xiao, J., Liu, Z.,
  Peng, L., Yan, S., Jin, D.: {DeepStationing: Thoracic Lymph Node Station
  Parsing in CT Scans using Anatomical Context Encoding and Key Organ
  Auto-Search}. In: MICCAI (2021)

\bibitem{Heinrich2012DEEDS}
Heinrich, M.P., Jenkinson, M., Brady, S.M., Schnabel, J.A.: {Globally optimal
  deformable registration on a minimum spanning tree using dense displacement
  sampling}. In: MICCAI. vol. 7512 LNCS, pp. 115--122 (2012).
  \doi{10.1007/978-3-642-33454-2_15}, \url{http://users.ox.ac.uk/{~}shil3388}

\bibitem{Isensee2021nnunet}
Isensee, F., Jaeger, P.F., Kohl, S.A., Petersen, J., Maier-Hein, K.H.:
  {nnU-Net: a self-configuring method for deep learning-based biomedical image
  segmentation}. Nature Methods  \textbf{18}(2),  203--211 (2021)

\bibitem{Iuga2021automated}
Iuga, A.I., Carolus, H., H{\"{o}}ink, A.J., Brosch, T., Klinder, T., Maintz,
  D., Persigehl, T., Bae{\ss}ler, B., P{\"{u}}sken, M.: {Automated detection
  and segmentation of thoracic lymph nodes from CT using 3D foveal fully
  convolutional neural networks}. BMC Medical Imaging  \textbf{21}(1), ~69
  (2021)

\bibitem{Liu2016station}
Liu, J., Hoffman, J., Zhao, J., Yao, J., Lu, L., Kim, L., Turkbey, E.B.,
  Summers, R.M.: {Mediastinal lymph node detection and station mapping on chest
  CT using spatial priors and random forest}. Medical Physics  \textbf{43}(7),
  4362--4374 (2016). \doi{10.1118/1.4954009}

\bibitem{Liu2020radam}
Liu, L., Jiang, H., He, P., Chen, W., Liu, X., Gao, J., Han, J.: {On the
  Variance of the Adaptive Learning Rate and Beyond}. In: ICLR (2020)

\bibitem{Mao2014current}
Mao, Y., Hedgire, S., Harisinghani, M.: {Radiologic Assessment of Lymph Nodes
  in Oncologic Patients}. Current Radiology Reports  \textbf{2}(2),  1--13 (feb
  2014)

\bibitem{Oda2018dense}
Oda, H., Bhatia, K.K., Roth, H.R., Oda, M., Kitasaka, T., Iwano, S., Homma, H.,
  Takabatake, H., Mori, M., Natori, H., Schnabel, J.A., Mori, K.: {Dense
  volumetric detection and segmentation of mediastinal lymph nodes in chest CT
  images}. In: Mori, K., Petrick, N. (eds.) SPIE. vol. 10575, p.~1. SPIE (Feb
  2018)

\bibitem{Perez-Garcia2020torchio}
P{\'{e}}rez-Garc{\'{i}}a, F., Sparks, R., Ourselin, S.: {TorchIO: a Python
  library for efficient loading, preprocessing, augmentation and patch-based
  sampling of medical images in deep learning}. Tech. rep. (2020),
  \url{https://github.com/fepegar/torchio}

\bibitem{Ren2015faster}
Ren, S., He, K., Girshick, R., Sun, J.: {Faster r-cnn: Towards real-time object
  detection with region proposal networks}. In: NIPS. pp. 91--99 (2015).
  \doi{10.1109/TPAMI.2016.2577031}

\bibitem{Roth201425D}
Roth, H.R., Lu, L., Seff, A., Cherry, K.M., Hoffman, J., Wang, S., Liu, J.,
  Turkbey, E., Summers, R.M.: {A New 2.5D Representation for Lymph Node
  Detection using Random Sets of Deep Convolutional Neural Network
  Observations}. In: MICCAI (2014)

\bibitem{Sharma2004pattern}
Sharma, A., Fidias, P., Hayman, L.A., Loomis, S.L., Taber, K.H., Aquino, S.L.:
  {Patterns of lymphadenopathy in thoracic malignancies}. Radiographics
  \textbf{24}(2),  419--434 (2004). \doi{10.1148/rg.242035075}

\bibitem{Smithuis2010}
Smithuis, R.: {Mediastinum Lymph Node Map} (2010),
  \url{https://radiologyassistant.nl/chest/mediastinum/mediastinum-lymph-node-map}

\bibitem{Tian2019FCOS}
Tian, Z., Shen, C., Chen, H., He, T.: {FCOS: Fully Convolutional One-Stage
  Object Detection}. In: ICCV (2019)

\bibitem{Yan2020LENS}
Yan, K., Cai, J., Zheng, Y., Harrison, A.P., Jin, D., Tang, Y.B., Tang, Y.X.,
  Huang, L., Xiao, J., Lu, L.: {Learning from Multiple Datasets with
  Heterogeneous and Partial Labels for Universal Lesion Detection in CT}. IEEE
  Trans. Med. Imaging  \textbf{2020}, ~1 (sep 2020)

\bibitem{Yan2019MULAN}
Yan, K., Tang, Y., Peng, Y., Sandfort, V., Bagheri, M., Lu, Z., Summers, R.M.:
  {MULAN: Multitask Universal Lesion Analysis Network for Joint Lesion
  Detection, Tagging, and Segmentation}. In: MICCAI. vol. 11769 LNCS, pp.
  194--202 (2019)

\bibitem{Zhu2020gating}
Zhu, Z., Jin, D., Yan, K., Ho, T.Y., Ye, X., Guo, D., Chao, C.H., Xiao, J.,
  Yuille, A., Lu, L.: {Lymph Node Gross Tumor Volume Detection and Segmentation
  via Distance-Based Gating Using 3D CT/PET Imaging in Radiotherapy}. In:
  MICCAI. pp. 753--762 (2020)

\end{thebibliography}

\end{document}